\documentclass[10pt,twocolumn]{article}

\usepackage[utf8]{inputenc} 
\usepackage{longtable}
\usepackage{graphicx}
\usepackage{epstopdf}
\usepackage[fleqn]{amsmath}
\usepackage{bm}
\usepackage{amsmath}
\usepackage{amssymb}
\usepackage{graphicx}
\usepackage{algorithm}
\usepackage{algorithmic}
\usepackage{booktabs}
\usepackage{url}
\usepackage{booktabs}
\usepackage{multirow}

\newtheorem{theorem}{Theorem}[section]
\newtheorem{lemma}{Lemma}[section]

\newcommand{\ro}{{\textnormal{o}}}

\newcommand{\0}{{\bm{0}}}
\newcommand{\1}{{\bm{1}}}

\newcommand{\vb}{{\bm{b}}}

\newcommand{\vd}{{\bm{d}}}

\newcommand{\ve}{{\bm{e}}}

\newcommand{\vh}{{\bm{h}}}

\newcommand{\x}{{\bm{x}}}

\newcommand{\y}{{\bm{y}}}

\newcommand{\vA}{{\bm{A}}}

\newcommand{\cC}{{\mathcal{C}}}

\newcommand{\vF}{{\bm{F}}}
\newcommand{\cF}{{\mathcal{F}}}
\newcommand{\vG}{{\bm{G}}}

\newcommand{\vI}{{\bm{I}}}

\newcommand{\bN}{{\mathbb{N}}}

\newcommand{\vO}{{\bm{O}}}

\newcommand{\bR}{{\mathbb{R}}}

\newcommand{\bS}{{\mathbb{S}}}
\newcommand{\cS}{{\mathcal{S}}}

\newcommand{\cW}{{\mathcal{W}}}
\newcommand{\vW}{{\bm{W}}}

\newcommand{\cX}{{\mathcal{X}}}

\newcommand{\valph}{{\bm{\alpha}}}

\newcommand{\vgam}{{\bm{\gamma}}}

\newcommand{\vxi}{{\bm{\xi}}}

\newcommand{\vPhi}{{\bm{\Phi}}}
\newcommand{\vSig}{{\bm{\Sigma}}}

\newcommand{\vThet}{{\bm{\Theta}}}

\DeclareMathOperator{\logm}{\textbf{logm}}

\newcommand{\argmax}{\mathop{\textrm{argmax}}\limits}
\newcommand{\argmin}{\mathop{\textrm{argmin}}\limits}

\newenvironment{tsaligned}{\begin{equation}\begin{aligned}}{\end{aligned}\end{equation}}

\usepackage[
top=15mm,
left=15mm,
right=15mm,
bottom=15mm,
footskip=5mm,
]{geometry}

\usepackage[theoremfont]{newtxtext}
\usepackage{newtxmath}

\newcommand{\constbregyo}{{B}}

\newcommand{\tslong}[1]{{#1}}
\newcommand{\tsshort}[1]{{}}

\begin{document}

\twocolumn[{%
\begin{center}
  {\Large
    Threshold Auto-Tuning Metric Learning}
\\
\vspace{1cm}
       {\large
         Yuya Onuma${}^{\dagger}$,
         Rachelle Rivero${}^{\dagger}$, 
         Tsuyoshi Kato${}^{\dagger,\ddagger,*}$}
\\
\vspace{1cm}
\begin{tabular}{lp{0.7\textwidth}}
${}^\dagger$ & 
Graduate School of Science and Technology, Gunma University, 
Kiryu-shi, Gunma, 376--8515, Japan.  
\\
${}^\ddagger$ &
Center for Research on Adoption of NextGen Transportation Systems, 
Gunma University, 
4--2 Aramaki-cho, Maebashi, Gunma 371--8510, Japan.
\end{tabular}
\end{center}
}]
\begin{abstract}
  It has been reported repeatedly that discriminative learning of distance metric boosts the pattern recognition performance. A weak point of ITML-based methods is that the distance threshold for similarity/dissimilarity constraints must be determined manually and it is sensitive to generalization performance, although the ITML-based methods enjoy an advantage that the Bregman projection framework can be applied for optimization of distance metric. In this paper, we present a new formulation of metric learning algorithm in which the distance threshold is optimized together. Since the optimization is still in the Bregman projection framework, the Dykstra algorithm can be applied for optimization. A nonlinear equation has to be solved to project the solution onto a half-space in each iteration. Na\"{i}ve method takes $O(LMn^{3})$ computational time to solve the nonlinear equation. In this study, an efficient technique that can solve the nonlinear equation in $O(Mn^{3})$ has been discovered. We have proved that the root exists and is unique. We empirically show that the accuracy of pattern recognition for the proposed metric learning algorithm is comparable to the existing metric learning methods, yet the distance threshold is automatically tuned for the proposed metric learning algorithm.
\end{abstract}

\section{Introduction}
Many reports have described so far that discriminative learning of distance metric on a feature space boosts the classification performance~\cite{Bellet13survey,Kulis13metric,KatTakOma13a}.
Recently, the research interests in distance metric learning are expanded to the application to non-vectorial data~\cite{Bellet13survey,Emms12stochastic,KatNag10a}.
In this paper, it is supposed that the object to be analyzed, $\x\in\cX$, where $\cX$ is the input space, is represented with a set of $M$ matrices
$\left(\vPhi_{m}(\x)\right)_{m=1}^{M}\in\cF$, where the $m$-th matrix $\vPhi_{m}(\x)$ has the size of $n_{m}\times n'_{m}$, 
$\vPhi_{m}(\cdot):\cX\to\bR^{n_{m}\times n'_{m}}$ is the $m$-th feature extractor,
and $\cF$ is the direct product of $M$ matrix sets
(i.e., $\cF:=(n_{1}\times n'_{1})\times\dots\times(n_{M}\times n'_{M})$).  
In this study, the following parameterized distance function
$D_{\vPhi}(\cdot,\cdot;\cW):\cX\times\cX\to\bR$ is discussed:
\begin{multline}\label{eq:D-Phi-def}
  D_{\vPhi}(\x_{1},\x_{2};\cW)
:=
\frac{1}{M}\sum_{m=1}^{M}
\Big<\vW_{m}, \\
\left(\vPhi_{m}(\x_{1})-\vPhi_{m}(\x_{2})\right) 
\left(\vPhi_{m}(\x_{1})-\vPhi_{m}(\x_{2})\right)^\top
\Big> 
\end{multline}
where $\x_{1},\x_{2}\in\cX$ and
$\cW := ( \vW_{1},\dots,\vW_{M} )$
is the parameter set of the distance function
$D_{\vPhi}(\x_{1},\x_{2};\cW)$ consisting of
$M$ strictly positive definite matrices
$\vW_{1}\in\bS_{++}^{n_{1}}$, $\dots$, $\vW_{M}\in\bS_{++}^{n_{M}}$,
where $\bS_{++}^{n}$ is the set of $n\times n$ strictly
positive definite matrices. 
Let $n=\max_{m}n_{m}$.
This function involves a broad class of parametric distances.
For example, in the setting of $M=1$ and $\vPhi_{1}:\cX\to\bR^{n}$,
$D_{\vPhi}(\x_{1},\x_{2};\cW)$ is the standard Mahalanobis distance
function between $n$-dimensional vectors
with a strictly positive definite Mahalanobis matrix $\vW_{1}$.
Recently, several works~\cite{MatRelSesKat16a,ZhiwuHuang15a} discuss distance metric learning for covariance descriptors, with the setting that $M=1$ and that $\vPhi_{m}(\x)$ is a covariance descriptor or its spectral variant.
The function $D_{\vPhi}(\x_{1},\x_{2};\cW)$ can be a distance among tensors by setting $\cX$ to an $M$-mode tensor space and by defining $\vPhi_{m}(\x)$ as the $m$-mode flattening of the tensor $\x\in\cX$.

Among the many metric learning methods for vectorial data, two of them~\cite{Bellet13survey,Kulis13metric}, \emph{LMNN} (large margin nearest neighbor)~\cite{WeiSau09a} and \emph{ITML} (information theoretic metric learning)~\cite{DavKulJaiSraDhi07a}, are especially popular; and many of their variants have been developed~\cite{Parameswaran10nips,Kedem12nonlinear,MatRelSesKat16a,ZhiwuHuang15a}.
LMNN employs \emph{relative distance constraints}, each of which demands that the distance between examples in different categories is greater than the distance between examples in the same category. 
Mahalanobis matrix is regularized with square Frobenius norm. 
Most methods in the family of LMNN contain no term that keeps the Mahalanobis matrix positive definite, and thereby the Mahalanobis matrix needs to be projected onto the positive definite cone to ensure the matrix to be positive definite. 

On the other hand, ITML is formulated with \emph{similarity/dissimilarity constraints} requiring that each pair in the same category should have a distance below a threshold, and that each pair in different categories should have a distance over a threshold.
The regularizer is the \emph{LogDet divergence} which is a \emph{Bregman divergence}~\cite{Bregman67a,RelNagKat16a,KatTakOma13a}. 
The LogDet divergence contains a term of $-\text{logdet}(\vW)$, which serves as a barrier function to ensure the positive definiteness of the Mahalanobis matrix~$\vW$. 
The objective function consists of a Bregman divergence from a constant point which allows use of \emph{Dykstra algorithm}~\cite{Censor98,MatRelSesKat16a} for optimization.

To use ITML and its variants, the threshold for the distance within same category and the distance between different categories must be manually determined in advance. The thresholds are referred to as the \emph{distance threshold} hereafter. 
The authors empirically found that the distance thresholds are sensitive to the pattern recognition performance.
It is expected that the usability would be improved if the distance threshold could be adjusted automatically.

In this study, we have developed a new metric learning algorithm, named \emph{Threshold Auto-Tuning Metric Learning} (TATML), based on Dykstra algorithm for determining the parameters $\cW := ( \vW_{1},\dots,\vW_{M} )$ for distance function \eqref{eq:D-Phi-def}. TATML does not deviate from the framework of the Dykstra algorithm which has many favorable properties, yet the distance thresholds can be adapted automatically within the unified Dykstra framework. 

This work is related to Matsuzawa et al.~\cite{MatRelSesKat16a}, whose study has developed a metric learning for a single covariance descriptor within the Dykstra framework. 
The region feasible to a distance constraint can be expressed as a \emph{half-space}.
Hence, if $K$ distance constraints are given, the metric learning task reduces to a problem of Bregman projection onto the intersection of $K$ half-spaces.
Dykstra algorithm finds the projection iteratively by projecting a current solution onto a single half-space randomly chosen at each iteration.
For metric learning of vectorial data, the projection onto a single half-space can be found in $O(n^{2})$ computation. 
In case of covariance descriptors, a nonlinear equation has to be solved to find the projection onto a half-space. Assessment of the value of both sides takes $O(n^{3})$ computation.
If it is $L$ times to assess the nonlinear equation in some numerical method such as Newton's method, $O(L n^{3})$ computation is required in a na\"{i}ve fashion to find a projection.
Matsuzawa et al.~\cite{MatRelSesKat16a} have found a computational trick that can find an exact projection within $O(n^{3})$ computation.

The difference of this study from \cite{MatRelSesKat16a} is that the distance function can contain multiple Mahalanobis matrices.
In this study, the authors found that projection onto a single half-space can be found efficiently even if multiple Mahalanobis matrices are contained in the distance function. 
A na\"{i}ve method takes $O(LMn^{3})$ computation for finding the projection, while the new algorithm needs only $O(Mn^{3})$ computation.

The contributions of this paper are summarized as follows.
\begin{itemize}
\item
  We present a new formulation for metric learning, named TATML, in which the distance thresholds are optimized together. Since the optimization is still in the Bregman projection framework, the Dykstra algorithm can be applied for optimization. 
\item
  A nonlinear equation has to be solved to project the solution onto a half-space in each iteration. Na\"{i}ve method takes $O(LMn^{3})$ computational time to solve the nonlinear equation. In this study, an efficient technique that can solve the nonlinear equation in $O(Mn^{3})$ has been discovered. We have proved that the root exists and is unique.
\item
Experimental results demonstrate that the accuracy of pattern recognition for TATML is comparable to the existing metric learning methods, yet the distance thresholds are automatically adapted for the proposed metric learning algorithm. 
\end{itemize}
%
\section{Formulation of Learning Problems}
\subsection{Bregman projection problem}
The \emph{Bregman divergence} is defined by
\begin{align}\label{eq:BD-def}
\text{BD}(\vThet,\vThet_{0}\,;\,\varphi)
= \varphi(\vThet) - \varphi(\vThet_{0}) - 
\left< \nabla \varphi(\vThet_{0}), \vThet-\vThet_{0} \right>, 
\end{align}
for the \emph{seed function}
$\varphi: \text{dom}(\varphi)\to \bR$
that is of Legendre type~\cite{rockafellar70convex}.
Note that $\vThet$ and $\vThet_{0}$ are not limited to
a vector, but can also be a matrix or a tuple of multiple matrices.
The domain of $\varphi$ is denoted by $\text{dom}(\varphi)$. 
The inner product
$\left<\cdot,\cdot\right>$ in \eqref{eq:BD-def}
is defined as the sum of
the products of each corresponding entries in
$\vThet$ and $\vThet_{0}$.  

For example,
for all $\x\in\cX$, 
letting $\varphi: \cX\to\bR$ be defined by 
\begin{equation}
  \varphi_{\text{maha}}\left(\x\right)
  :=
  \frac{1}{M}
  \sum_{m=1}^{M}
  \left<\vW_{m}, \vPhi_{m}(\x)\vPhi_{m}(\x)^\top
  \right>
\end{equation}
yields the parametric distance function~\eqref{eq:D-Phi-def}
as, $\forall \x_{1}, \forall \x_{2}\in\cX$, 
\begin{align}
  \text{BD}
  \left( \left(\vPhi_{m}(\x_{1})\right)_{m=1}^{M},
  \left(\vPhi_{m}(\x_{2})\right)_{m=1}^{M}\,;\,
  \varphi\right)
  =
  D_{\vPhi}(\x_{1},\x_{2};\cW).  
\end{align}
The divergence $D_{\vPhi}(\cdot,\cdot;\cW)$ satisfies the distance axiom if the mapping $\x\mapsto \left(\vPhi_{m}(\x)\right)_{m=1}^{M}$ is \emph{injective}.  Hence, the Bregman divergence is not a distance function in general
because it fails the symmetry condition. 

In ITML, the metric learning task is formulated with a Bregman projection problem. 
The Bregman projection from a point $\vThet_{0}$ onto a set $\cS$ is defined as the point in $\cS$ that has the minimal Bregman divergence from the point $\vThet_{0}$. Namely, it is
\begin{align}
  \argmin_{\vThet\in\cS} \text{BD}(\vThet,\vThet_{0}\,;\,\varphi). 
\end{align}
Bregman divergence is a strictly convex function, thereby making the Bregman projection unique. 

\subsection{Problems with manually determined distance thresholds}
\label{ss:prob-manual-b0}
To perform supervised learning in determining the value of the set of Mahalanobis matrices $\cW$, suppose we are given $\ell$ labeled data points $\x_{1}$,$\dots$,$\x_{\ell}\in\cX$.
The metric learning problem is formulated to improve the classification performance of the nearest neighbor classifier in the multi-category classification scenario.
In metric learning using similarity/dissimilarity constraints, $K$ example pairs are selected from $\ell$ examples, so that the first $K_{+}$ pairs $(i_{k},j_{k})\in\bN^{2}$ $(k=1,\dots,K_{+})$ belong to same category and the rest of $K_{-}(:=K-K_{+})$ pairs belong to different categories.
In some work~(e.g. \cite{WeiSau09a}), the pairs of smaller distance are chosen instead of random selection.

In the similarity/dissimilarity constraints, it is demanded that 
two examples in the same category are at most a distance of an upper bound $b_{\text{ub}}$ apart and two examples in different categories are at least a distance of a lower bound $b_{\text{lb}}$ apart. By setting $b_{k}=b_{\text{ub}}$ for $k=1,\dots,K_{+}$ and $b_{k}= b_{\text{lb}}$ for $k=K_{+}+1,\dots,K$, these requirements can be translated to the following $K$ constraints: 
\begin{tsaligned}
  \begin{cases}
    b_{k}-D_{\vPhi}(\x_{i_{k}},\x_{j_{k}};\cW) \ge 0, \quad
    &\text{for }k=1,\dots,K_{+},
    \\
    b_{k}-D_{\vPhi}(\x_{i_{k}},\x_{j_{k}};\cW) \le 0, \quad
    &\text{for }k=K_{+}+1,\dots,K. 
  \end{cases}
\end{tsaligned}
However, there might not exist $\cW$ that fulfills all the $K$ constraints. To ensure the existence of a solution, the constraints are softened by introducing slack variables
$\vxi = \left[\xi_{1},\dots,\xi_{K}\right]^\top$ as
\begin{align}\label{eq:kth-con}
  \begin{cases}
  \xi_{k}-D_{\vPhi}(\x_{i_{k}},\x_{j_{k}};\cW)
    \ge 0\quad\text{for }k=1,\dots,K_{+}, 
    \\
  \xi_{k}-D_{\vPhi}(\x_{i_{k}},\x_{j_{k}};\cW)    
    \le 0\quad\text{for }k=K_{+}+1,\dots,K.   
  \end{cases}
\end{align}
In ITML~\cite{DavKulJaiSraDhi07a} and its variants~\cite{MatRelSesKat16a,ZhiwuHuang15a},
a penalty defined by Bregman divergence $\text{BD}(\vxi,\vb\,;\,\varphi_{\ell})$
is introduced to find the model parameter that yields a small penalty.
Therein, $\varphi_{\ell}$ is the seed function of the penalty function.
In ITML~\cite{DavKulJaiSraDhi07a}, Kullback-Leibler divergence is employed
as the penalty for violation of similarity/dissimilarity constraints.

In addition to the penalty, ITML-based methods introduce another Bregman divergence for regularization of parameters, and employ the objective function that is the sum of the penalty and the regularization term:
\begin{align}\label{eq:obj-old-itml}
  P_{\text{o}}(\cW,\vxi) :=
  \text{BD}(\cW,\cW_{0}\,;\,\varphi_{\text{r}})
  +
  c \text{BD}(\vxi,\vb\,;\,\varphi_{\ell}), 
\end{align}
where both $c>0$ and $\cW_{0}$ are constants, and typically
we can set $\cW_{0}:=(\vI_{n_{1}},\dots,\vI_{n_{M}})$.
Then, the resultant distance function is not too apart from the Euclidean distance and yields small violations of similarity/dissimilarity constraints.
In this study, following the studies of \cite{DavKulJaiSraDhi07a,MatRelSesKat16a,ZhiwuHuang15a}, 
\begin{align}
  \varphi_{\text{r}}(\cW) := -\sum_{m=1}^{M}\text{logdet}(\vW_{m})
\end{align}
is employed as a seed function for regularization.
An advantage of this seed function is that the resultant $M$ Mahalanobis matrices are kept strictly positive definite.

We shall show that the optimization problem, which is to minimize the objective function~\eqref{eq:obj-old-itml} subject to $K$ constraints~\eqref{eq:kth-con}, is a Bregman projection problem. 
For $m=1,\dots,M$, $k=1,\dots,K$, we introduce
positive semidefinite matrices
$\vA_{m,k}:= \left(\vPhi_{m}(\x_{i_{k}})-\vPhi_{m}(\x_{j_{k}})\right) 
\left(\vPhi_{m}(\x_{i_{k}})-\vPhi_{m}(\x_{j_{k}})\right)^\top$. 
The set of $(\cW,\vxi)$ satisfying each of $K$ constraints
is expressed with a half-space: for $k= 1,\dots,K_{+}$
\begin{align}
  \cC_{k} :=
  \left\{
  (\cW,\vxi) \,\middle|\,
  \frac{1}{M}\sum_{m=1}^{M}\left<\vA_{m,k},\vW_{m}\right> \le \xi_{k}
  \right\}, 
\end{align}
and for $k=K_{+}+1,\dots,K$
\begin{align}
  \cC_{k} :=
  \left\{
  (\cW,\vxi) \,\middle|\,
  \frac{1}{M}\sum_{m=1}^{M}\left<\vA_{m,k},\vW_{m}\right> \ge \xi_{k}
  \right\}. 
\end{align}
By defining a seed function  as
\begin{align}
  \varphi_{\text{o,tot}}(\cW,\vxi) :=
  \varphi_{\text{r}}(\cW) + c \varphi_{\ell}(\vxi), 
\end{align}
it can be observed that the objective function
is the Bregman divergence generated from the seed
function $\varphi_{\text{o,tot}}(\cdot)$. Namely,
it is established that
\begin{align}\label{eq:Po-is-BD-tot}
 P_{\text{o}}(\cW,\vxi) = \text{BD}((\cW,\vxi),(\cW_{0},\vb)\,;\,\varphi_{\text{o,tot}}). 
\end{align}
Hence, it has been shown that the metric learning problem of
minimizing the objective function~\eqref{eq:obj-old-itml}
subject to $K$ constraints \eqref{eq:kth-con}
is reduced to the
problem of finding the Bregman projection from
a point $(\cW_{0},\vb)$ onto
\begin{align}
  \bigcap_{k=1}^{K}\cC_{k}. 
\end{align}
Thus, since the metric learning problem described above is a Bregman projection problem, the optimal solution can be found by Dykstra method. 
However, from our preliminary experiments, we have found that the value of $\vb$ is sensitive to the generalization performance for pattern recognition.
This degenerates the usability due to manual tuning of the hyper parameter $\vb$.

\subsection{Problems with automatically tuned distance thresholds}
In the learning problem mentioned previously, the distance thresholds are treated as a constant. Here, in order to alleviate the cumbersome step of adjusting the distance thresholds manually, each entry in the constant vector $\vb\in\bR^{K}$ is changed to a function of a single scalar $b_{0}\in\bR$ as
\begin{align}
  b_{k}(b_{0}) = 
  \begin{cases}
    b_{0}/2  &\qquad
    \text{for } k= 1,\dots,K_{+}
    \\
    2b_{0} &\qquad
    \text{for } k=K_{+}+1,\dots,K.  
  \end{cases}
\end{align}
We consider simultaneous optimization of $(\cW,\vxi)$ and $b_{0}$, hereinafter.  
By letting $\vgam =
\left[\gamma_{1},\dots,\gamma_{K}\right]^\top :=
\left[2^{-1}\1_{K_{+}}^\top, 2\1_{K_{-}}^\top\right]^\top$,
we can write $\vb = b_{0}\vgam$.
Not only for $\cW$, we introduce a regularization term 
\begin{align}
  \text{BD}(b_{0},\mu_{0};\varphi_{\ell0})
\end{align}
for a new variable $b_{0}$ and a constant $\mu_{0}$. 
Then, the objective function of this metric learning problem
can be expressed as
\begin{multline}\label{eq:P-b0-def}
  P_{\text{b0}}(\cW,\vxi,b_{0}) :=
  \text{BD}(\cW,\cW_{0}\,;\,\varphi_{\text{r}})
  \\
  +
  c_{0} \text{BD}(b_{0},\mu_{0};\varphi_{\ell0})
  +
  c \text{BD}(\vxi,b_{0}\vgam\,;\,\varphi_{\ell}), 
\end{multline}
where $c_{0} >0$ is a constant
for the regularization term of $b_{0}$.
The problem of simultaneous optimization of $\cW$ and $b_{0}$ has more hyperparameters than the above-mentioned problem of minimizing~\eqref{eq:obj-old-itml}, although no hyperparameters are sensitive to the generalization performance.
However, one may notice that minimizing~\eqref{eq:P-b0-def} is more intractable.
On the one hand,
the function $P_{\text{o}}(\cdot,\cdot)$ is a Bregman divergence from a constant point
as discussed in Sect.~\ref{ss:prob-manual-b0}. 
Therefore, minimizing $P_{\text{o}}(\cdot,\cdot)$ is a Bregman projection problem. 
On the other hand, 
the function $P_{\text{b0}}(\cdot,\cdot,\cdot)$ is no more a Bregman divergence
from a constant point nor a convex function, which makes optimization difficult. 

The main theoretical finding of this study is to discover a special setting in which minimizing the function $P_{\text{b0}}(\cW,\vxi,b_{0})$ with respect to $\cW$, $\vxi$, and $b_{0}$ subject to $K$ constraints is reduced to a Bregman projection problem, yet the optimization is an intractable non-convex problem in a general setting.
The minimization problem we shall discuss is
\begin{tsaligned}\label{eq:prob-Pb0}
  \text{min}\quad&
  P_{\text{b0}}(\cW,\vxi,b_{0})
  \quad
  \text{wrt}
  \quad
  \cW, \vxi, b_{0},
  \\
  \text{subject to}\quad&
  (\cW,\vxi) \in \cC_{k}
  \quad
  \text{for}
  \quad
  k=1,\dots,K. 
\end{tsaligned}
which can be equivalently rewritten as
\begin{tsaligned}
  \text{min}\quad&
  P_{\text{b}}(\cW,\vxi)
  \quad
  \text{wrt}
  \quad
  (\cW,\vxi) \in \bigcap_{k=1}^{K}\cC_{k}, 
\end{tsaligned}
where we have defined
\begin{tsaligned}
  P_{\text{b}}(\cW,\vxi)
  :=
  \min_{b_{0}\in\text{dom}(\varphi_{\ell 0})}
  P_{\text{b0}}(\cW,\vxi,b_{0}). 
\end{tsaligned}
In this study, we have discovered that the objective function
$P_{\text{b}}(\cdot,\cdot)$
is equal to a Bregman divergence up to a constant, in the setting of
\begin{align}\label{eq:seed-ell0-and-ell-def}
  \varphi_{\ell 0}(b_{0}) &:= \frac{1}{2}b_{0}^{2},
  &
  \varphi_{\ell}(\vxi) &:=
  \frac{1}{2}\lVert\vxi\rVert^{2}. 
\end{align}
\begin{lemma}
  \label{lem:biasoptloss-is-bd}
  {\it
  If $c_{0}>0$ and
  $\varphi_{\ell 0}$ and $\varphi_{\ell}$ are
  defined as \eqref{eq:seed-ell0-and-ell-def}, 
  then there exist
  $\vG\in\bS_{++}^{K}$, 
  $\vxi_{0}\in\bR^{K}$, and
  $\constbregyo\in\bR$ such that
  $\forall \cW$, 
  $\forall \vxi$, 
  \begin{align}\label{eq:Pb-is-bd}
    P_{\text{b}}(\cW,\vxi) + \constbregyo
    =
    \text{BD}((\cW,\vxi),(\cW_{0},\vxi_{0})\,;\,\varphi_{\text{tot}}),  
  \end{align}
  where the seed function of the Bregman divergence
  in RHS is given by
  \begin{align}\label{eq:seed-tot-def}
    \varphi_{\text{tot}}(\cW,\vxi) :=
    \varphi_{\text{r}}(\cW) + 
    \frac{1}{2}\left<\vxi,\vG\vxi\right>. 
  \end{align}
  }
\end{lemma}

\tslong{See Sect.~\ref{ss:proof-lem:biasoptloss-is-bd}
  for proof of Lemma~\ref{lem:biasoptloss-is-bd}.} 
Actually, the equality~\eqref{eq:Pb-is-bd} follows
by setting
\begin{equation}\label{eq:G-xi0-def-for-biasoptloss}
\begin{aligned}
  \vG &= \left(\vI + \mu_{3}\vgam\vgam^\top\right)c, 
  \\
  \vxi_{0} &:=
  \left(
  c \mu_{2} +
  \left(
  c_{0}\mu_{1}\mu_{0} - c\mu_{2}\right)
  \mu_{1}\lVert\vgam\rVert^{2}
  \right)
  \vG^{-1}\vgam, 
\end{aligned}
\end{equation}
and
\begin{align}
  \constbregyo &:= 
  \frac{1}{2}
  \left<
  \vG\vxi_{0},\vxi_{0}\right>
  -\frac{1}{2}
  c \mu_{2}^{2}\lVert\vgam\rVert^{2}
  -\frac{1}{2}
  c_{0}\mu_{1}^{2}\mu_{0}^{2}\lVert\vgam\rVert^{4}, 
\end{align}
where
$\mu_{1} := c/(c\lVert\vgam\rVert^{2}+c_{0})$,
$\mu_{2} := c_{0}\mu_{0}/(c\lVert\vgam\rVert^{2}+c_{0})$,
and
$\mu_{3} := \mu_{1}^{2}\left(c_{0}/c+\lVert\vgam\rVert^{2}\right) -2\mu_{1}$. 
This also implies that the regularization constant $c_{0}$ must be positive because $\vG$ would be singular if $c_{0}=0$ \tslong{(See Lemma~\ref{lem:G-is-pd})}.
Thus, we have obtained the following result. 
\begin{theorem}
  {\it
  Problem~\eqref{eq:prob-Pb0} is a Bregman projection problem, 
  provided that
  two seed functions, $\varphi_{\ell 0}$ and $\varphi_{\ell}$,
  are defined as   \eqref{eq:seed-ell0-and-ell-def}
  and that $c_{0}>0$. }
\end{theorem}

The proposed algorithm TATML employs the problem~\eqref{eq:prob-Pb0} in order to determine the distance metric $\cW$. 

\section{Optimization Algorithm for TATML}
In this section, an algorithm for solving a metric learning problem~\eqref{eq:prob-Pb0} is presented.
In the previous section, we have described that the problem~\eqref{eq:prob-Pb0} can be reduced to a Bregman projection from a constant point~$(\cW_{0},\vxi_{0})$ onto the intersection of $K$ half-spaces $\cC_{k}$. 
In TATML, the stochastic Dykstra algorithm~\cite{MatRelSesKat16a} is employed to solve this Bregman projection problem.  The stochastic Dykstra algorithm is an iterative method, and at $t$-th iteration the previous solution $(\cW_{t-1},\vxi_{t-1})$ is projected onto the boundary of the $k$-th half-space $\text{bd}(\cC_{k})$, where $(\cW_{t-1},\vxi_{t-1})$ is the solution obtained at $(t-1)$-th iteration.
Projection onto $\text{bd}(\cC_{k})$ is equivalent to the following mini-max problem $\text{SP}(\cW_{t-1},\vxi_{t-1})$:
\begin{tsaligned}\label{eq:minimax-biasitml}
  \text{SP}(\cW_{t-1},\vxi_{t-1}):\quad
  \max_{\bar{\delta}\in\bR}\min_{(\cW,\vxi)} Q_{t}(\cW,\vxi,\bar{\delta})
\end{tsaligned}
where
\begin{multline}\label{eq:Q-minimax-biasitml-def}
  Q_{t}(\cW,\vxi,\bar{\delta})
  :=
  \text{BD}((\cW,\vxi),(\cW_{t-1},\vxi_{t-1})\,;\,\varphi_{\text{tot}})
  \\
  +
  \left(
  \frac{1}{M}
  \sum_{m=1}^{M}
  \left<\vA_{m,k},\vW_{m}\right>
  -
  \left<\ve_{k},\vxi\right>
  \right)
  \bar{\delta}. 
\end{multline}
The Dykstra algorithm applied to the metric learning problem~\eqref{eq:prob-Pb0}
is given as follows.
\begin{itemize}
\item \textit{Step 1}: 
  $\y := \left[\1_{K_{+}}^\top, -\1_{K_{-}}^\top\right]^\top$;
  $\valph^{(0)} := \0_{K}$; 
\item \textit{Step 2}: For $k\in\{1,\dots,K\}$, set $\vh_{k}$ to
  the $k$-th column of $\vG^{-1}$;  
\item \textit{Iterate}: for $t := 1,2,\dots$
  \begin{itemize}
  \item \textit{Step 3}: Pick $k$ randomly from $\{1,\dots,K\}$;
  \item \textit{Step 4}:
    $\bar{\delta}_{t}:=\argmax_{\delta}\min_{(\cW,\vxi)} Q_{t}(\cW,\vxi,\delta)$;
  \item \textit{Step 5}: $\delta_{t}:=\max\left(y_{k}\bar{\delta}_{t},-\left<\ve_{k},\alpha^{(t-1)}\right>\right)$; \\
    \quad \qquad $\valph^{(t)}:=\valph^{(t-1)} + \delta_{t}\ve_{k}$;
    \, \\
    \quad \qquad $\vxi_{t} := \vxi_{t-1} + \delta_{t}y_{k}\vh_{k}$; 
  \item \textit{Step 6}: $\forall m\in\{1,\dots,M\}$, \\ \qquad\qquad
    $\vW_{t,m} := \left(\vW_{t-1,m}^{-1} + \frac{{\delta_{t}y_{k}}}{M}\vA_{m,k}\right)^{-1}$; \\ \quad \qquad $\cW_{t}:=(\vW_{t,1},\dots,\vW_{t,M});$ 
  \end{itemize}
\end{itemize}
Therein, in the description of this algorithm,
we have denoted the $m$-th Mahalanobis matrix at $t$-th iteration
by $\vW_{t,m}$.  As mentioned in Sect.~\ref{ss:prob-manual-b0}, 
the $m$-th matrix in $\cW_{0}$, denoted by
$\vW_{0,m}$, is the $n_{m}\times n_{m}$ identity matrix
in a typical setting.  
\subsection{Solution to $\text{SP}(\cW_{t-1},\vxi_{t-1})$}
In Step 4, the solution $(\cW_{t-1},\vxi_{t-1})$ is
projected onto $\text{bd}(\cC_{k})$
by solving the mini-max problem $\text{SP}(\cW_{t-1},\vxi_{t-1})$.
The solution to $\text{SP}(\cW_{t-1},\vxi_{t-1})$
satisfies the stationary condition of
$Q_{t}(\cdot,\cdot,\cdot)$
\begin{tsaligned}
  \nabla_{\bar{\delta}}Q_{t}&=0, &
  \nabla_{\vxi}Q_{t}&=\0_{K},&
  \forall m, \, \nabla_{\vW_{m}}Q_{t}&=\vO_{n_{m}} 
\end{tsaligned}
and the positive definiteness
\begin{tsaligned}
  \vW_{1}\succ \vO, \quad \dots, \quad
  \vW_{K}\succ \vO. 
\end{tsaligned}
From the stationary condition, we get
the following nonlinear equation of $\delta$:
\begin{multline}\label{eq:nonlineq}
  \left<\ve_{k},
  \vxi_{t-1}\right>
  +
  \bar{\delta} h_{k,k}
  \\
  =
  \frac{1}{M}  
  \sum_{m=1}^{M}\left<\vA_{m,k},
  \left(
  \vW_{t-1,m}^{-1}
  + \bar{\delta}\vA_{m,k}
  \right)^{-1}
  \right>, 
\end{multline}
where $h_{k,k}$ is the $k$-th diagonal entry
in $\vG^{-1}$. 

\paragraph*{Na\"{i}ve Projection:}
The nonlinear equation \eqref{eq:nonlineq} is not solved in a closed form. Hence, the use of some numerical methods such as Newton's method is required to find the root of the nonlinear equation.  
In a numerical method, the values of the both sides of \eqref{eq:nonlineq} have to be assessed at several values of $\bar{\delta}$.
Assume \eqref{eq:nonlineq} is assessed at $L$ values of $\bar{\delta}$.
Since RHS of \eqref{eq:nonlineq} contains $M$ inverse matrices and each needs $O(n^{3})$ computation, the time complexity of this approach to solve the nonlinear equation~\eqref{eq:nonlineq} is $O(LM n^{3})$. 

\paragraph*{Efficient Projection:}
In this study, we have found that exact projection onto a half-space can be obtained within $O(Mn^{3})$ computation by using a trick similar to the one presented in \cite{MatRelSesKat16a}. 
The time complexity $O(Mn^{3})$ comes from the fact that an $N(:=n_{1}+\dots+n_{M})$-dimensional vector $\vd:=\left[d_{1},\dots,d_{N}\right]^\top\in\bR_{++}^{N}$ satisfying
\begin{tsaligned}\label{eq:Ndim-dvec-def}
  \text{RHS of \eqref{eq:nonlineq}}
  =
  \text{tr}(\text{diag}(\vd + \bar{\delta}\1)^{-1})
\end{tsaligned}
can be found in $O(Mn^{3})$ computation.
Once $\vd$ is found, RHS of \eqref{eq:nonlineq} can
be assessed with $O(N)$ computation.
Hence, the nonlinear equation~\eqref{eq:nonlineq}
can be solved in $O(Mn^{3})$ computation
if $L\in O(n^{2})$.

\paragraph*{How to Compute $\vd$:}
Indeed, the $N$-dimensional vector satisfying 
\eqref{eq:Ndim-dvec-def} can be found as follows.
From each of $M$ strictly positive definite matrices
$\vA_{m,k}^{-1/2}\vW_{m,t-1}\vA_{m,k}^{-1/2}\in\bS_{++}^{n_{m}}$,
$n_{m}$ (not necessarily distinct)
eigenvalues can be obtained. The number of
eigenvalues from the $M$ matrices is $N$ in total.
Concatenating these $N$ eigenvalues yields the
vector $\vd$ satisfying \eqref{eq:Ndim-dvec-def}.
To ensure that each of $\vW_{m}$ is strictly
positive definite, $\bar{\delta}$ must be
found in the left-bounded open interval 
$(-d_{\text{min}},+\infty)$ where
$d_{\text{min}}:=\min_{h=1}^{N}d_{h}$.


\subsection{Computation of $\vh_{k}$}
In Step 2, $\vh_{k}\in\bR^{K}$ is defined
as the $k$-th column in the inverse of
a $K\times K$ matrix $\vG$.
If $\vG$ is computed using the standard algorithm
for inverse computation such as LU decomposition,
the time complexity for computing $\vh_{k}$
is $O(K^{3})$, which makes the entire algorithm
heavy if $K$ is large.
However, due to the special structure of $\vG$,
$\vh_{k}$ can be given analytically as
\begin{align}\label{eq:hk-order-K}
  \vh_{k} = \frac{1}{c}\ve_{k}
  - \frac{\mu_{3}\gamma_{k}}{c(1+\mu_{3}\lVert\vgam\rVert^{2})}\vgam. 
\end{align}
Using this equality, each of $\vh_{k}$ can be
obtained in $O(K)$. Hence, Step 2 takes only $O(K^{2})$
computation.

\section{Experiments}
In this section we report experimental results on the performance of our metric learning method, TATML. One advantage of TATML is automatic tuning of distance thresholds. The experimental results reveal that the pattern recognition performance of TATML is comparative to the existing metric learning methods, yet TATML adjusts the distance thresholds automatically. 

\subsection{Experimental settings}
We conducted experiments on three datasets, ETH-80~\cite{Leibe03}, Brodatz~\cite{Brodatz99}, and Virus~\cite{Kylberg11}. The number of images for each dataset are as follows: 3,280 (ETH-80); 112 (Brodatz); and 1,500 (Virus). For each dataset, almost a half of them were chosen at random for training, and the rest were for testing. We repeated this procedure five times, and the average of the five accuracies for multi-category classification was taken. In this section, all the generalization performances are reported using those average accuracies. A covariance descriptor was extracted from each image. The sample covariance matrix was obtained from local feature vectors, each of which was computed at a pixel. The pattern recognition was performed with the nearest neighbor classifier using the covariance descriptor in the distance metric space determined by the learning algorithm. The number of neighbors was set to three.


One of the settings used in the experiments was $M=1$, and
$\vPhi_{1}(\x):=\text{logm}(\vSig_{1}(\x))$
where $\text{logm}(\cdot):\bS^{n_{1}}_{++}\to\bR^{n_{1}\times n_{1}}$ is the function taking
the matrix logarithm of a strictly positive definite argument, and
$\vSig_{1}(\x)$ is the sample covariance matrix computed in the way described above from an image~$\x$.  A small number is added to diagonal entries in each sample covariance matrix to ensure these matrices in the effective domain of $\logm$. 

We also tested another covariance descriptor, denoted by $\vSig_{.5}(\cdot)$. We halved the scale of each image to consider two other settings. The one was $M=1$ and $\vPhi_{1}(\x):=\text{logm}(\vSig_{.5}(\x))$, and in the other, $M=2$ covariance descriptors were used with $\vPhi_{1}(\x):=\text{logm}(\vSig_{1}(\x))$ and
$\vPhi_{2}(\x):=\text{logm}(\vSig_{.5}(\x))$.
In total, the covariance descriptors used in the experiments were three types
summarized in Table~\ref{tab:three-featyps}. 
\begin{table}[t]
  \begin{center}
    \caption{Three feature types used in our experiments. \label{tab:three-featyps}}
    \begin{tabular}{ll}
      \toprule
      Feature Type
      &
      Description
      \\
      \midrule
      $M=2; (1,.5)$ &
      $(\text{logm}(\vSig_{1}(\x)),\text{logm}(\vSig_{.5}(\x)))$ 
      \\
      $M=1; (1)$ &
      $(\text{logm}(\vSig_{1}(\x)))$ 
      \\
      $M=1; (.5)$ &
      $(\text{logm}(\vSig_{.5}(\x)))$ 
      \\
      \bottomrule
    \end{tabular}
  \end{center}
\end{table}

Using the three types of features, we compared four methods, \textbf{TATML}, \textbf{Euc}, \textbf{Maz}, and \textbf{ITML}. TATML is the proposed metric learning algorithm that determines the value of the distance metric $\cW$. Euc fixes the Mahalanobis matrices to $\vW_{m}:=\vI_{n_{m}}$ for $m=1,\dots,M$, which is equivalent to use of Euclidean distance metric. Maz minimizes $P_{\ro}$ to learn the distance metric $\cW$. In case of $M=1$, Maz is equivalent to the method proposed by Matsuzawa et al.~\cite{MatRelSesKat16a}. ITML is a state-of-the-art metric learning method for vectors~\cite{DavKulJaiSraDhi07a}. In our experiments, the vectors were obtained by vectorizing $M$ covariance descriptors and concatenating the vectors to apply ITML. The dimensionality of the vectors was adjusted with the principal component analysis so that the number of degrees of freedom is almost the same as that of TATML. 

For each category, 10 constraints were imposed: five constraints were derived from two examples in the same category, and the rest were for different categories.
the two hyperparameters were fixed to $c_{0}=1$ and $\mu_{0}=1$.
The value of regularization parameter $c$ is selected from $\{10^{-2},10^{-1},10^0,10^1\}$, by using cross-validation within the training dataset.
For Maz and ITML, the optimal distance thresholds $(b_{\text{ub}}, b_{\text{lb}})$ were also searched exhaustively from a wide discrete range: $(b_{0}/2,2b_{0})$ for $b_{0}=2^{-2},2^{-1},2^{0},2^{1}$. 



\subsection{Comparisons to existing methods}
\begin{table}[t]
  \centering
  \caption{
    Generalization performance of pattern recognition.
    Three columns represent three different features,
    `$M=2$; $(1, .5)$', `$M=1$; $(1)$', `$M=1$; $(.5)$'.  
    The details of each feature is described in the main text.
    The font of the highest accuracy in each column is boldfaced. 
    Accuracies without significant difference from the highest accuracies are underlined. 
    \label{tab:mean-performance}
  }
  \newcommand{\singleto}[2]{\multicolumn{1}{#1}{#2}}
  \begin{tabular}{l}
    (a) ETH-80. \\
    \begin{tabular}{lccc}
      \toprule
      & \singleto{l}{$M=2$; $(1, .5)$}
      & \singleto{l}{$M=1$; $(1)$}
      & \singleto{l}{$M=1$; $(.5)$} \\
      \midrule
      TATML & \underline{95.71} & \underline{95.27} & \underline{95.36} \\
      Euc & 95.38 & 93.49 & 95.24 \\
      Maz & \underline{95.83} & \textbf{95.36} & \textbf{95.56} \\
      ITML & \textbf{96.07} & 93.93 & \underline{93.77} \\
      \bottomrule
    \end{tabular}
    \\
    \\
    (b) Brodatz Texture.
    \\
    \begin{tabular}{lccc}
      \toprule
      & \singleto{l}{$M=2$; $(1, .5)$}
      & \singleto{l}{$M=1$; $(1)$}
      & \singleto{l}{$M=1$; $(.5)$} \\
      \midrule
      TATML & \textbf{86.42} & \textbf{84.90} & \textbf{79.05} \\
      Euc & 85.88 & 84.53 & 78.39 \\
      Maz & \underline{86.21} & \underline{84.79} & \underline{78.85} \\
      ITML & 81.17 & 79.07 & 72.38 \\
      \bottomrule
    \end{tabular}
    \\ \\
    (c) Virus Texture.
    \\
    \begin{tabular}{lccc}
      \toprule
      & \singleto{l}{$M=2$; $(1, .5)$}
      & \singleto{l}{$M=1$; $(1)$}
      & \singleto{l}{$M=1$; $(.5)$} \\
      \midrule
      TATML & \underline{64.93} & \textbf{61.73} & \underline{50.67} \\
      Euc & 60.53 & 58.13 & 43.87 \\
      Maz & \textbf{66.80} & \underline{61.47} & \textbf{51.20} \\
      ITML & 47.40 & 59.73 & 41.20 \\
      \bottomrule
    \end{tabular}
  \end{tabular}
\end{table}
Table~\ref{tab:mean-performance} reports the accuracy of multi-category classification on three datasets: ETH-80, Brodatz, and Virus. For Brodatz dataset, the proposed method, TATML, achieved the highest accuracy for all feature types. For Virus dataset, the highest accuracies were observed by Maz when `$M=2;(1,.5)$' and `$M=1;(.5)$' were used, whereas TATML yields the best performance for `$M=1;(1)$'.
We performed the one-sample $t$-test to examine whether the highest accuracy in each column of Table~\ref{tab:mean-performance} was significantly higher than the other accuracies. The significance level was set to 5\%. It is noteworthy that TATML achieved best accuracies in many experimental settings, and, for all the other settings, the accuracies of TATML were not significantly lower from the highest accuracy. 

\subsection{Are hyperparameters sensitive?}
A shortcoming of the existing ITML-based methods is that the distance thresholds $(b_{\text{ub}}, b_{\text{lb}})$ have to be determined manually, although the generalization performance is highly dependent on the values of the distance thresholds. To illustrate the shortcoming, we conducted additional experiments with the distance thresholds fixed to particular four values. When $(b_{\text{ub}}, b_{\text{lb}})=(1/2,2)$, the accuracy was $95.88$ on ETH-80. In the other settings of $(b_{\text{ub}}, b_{\text{lb}})$, the one-sample $t$-test concluded that the accuracies were significantly lower than that of $(b_{\text{ub}}, b_{\text{lb}})=(1/2,2)$, suggesting that in the ITML-based methods the dependency on the hyperparameters $(b_{\text{ub}}, b_{\text{lb}})$ is not ignorable.

A remarkable characteristic of TATML is acquisition of ability to adapt the distance thresholds automatically to the training dataset, although two hyperparameters,
$c_{0}$, and $\mu_{0}$,
are introduced newly. We shall demonstrate how sensitive to the generalization performance the two hyper-parameters are. We varied the two hyper-parameters with
$c_{0}=0.25, 0.5, 1.0$, and $\mu_{0}=0.25, 0.5, 1.0, 2.0$ exhaustively, then all the combinations yielded $12 (=4\times 3)$ accuracies in total.
All 12 accuracies are exactly equal to $95.929$.

\section{Conclusions}
In this paper, we have presented a new metric learning algorithm, named TATML, that overcomes a shortcoming of existing ITML-based methods. The shortcoming is that the distance thresholds must be determined manually and it is sensitive to generalization performance, although the ITML-based methods enjoy
an advantage that learning the distance metric can be performed in the
Bregman projection framework. TATML optimizes the distance metric as well as the distance thresholds simultaneously. Since the optimization problem of TATML is still in the Bregman projection framework, the Dykstra algorithm can be exploited for optimization, in which a nonlinear equation has to be solved at each iteration. Na\"{i}ve method takes $O(LMn^{3})$ computational time to solve the nonlinear equation. In this study, we have found an efficient technique that can solve the nonlinear equation in $O(Mn^{3})$. We have theoretically proved the existence and uniqueness of the root of the nonlinear equation. We empirically show that the accuracy of pattern recognition for TATML is comparable to the existing metric learning methods, yet the distance thresholds are automatically tuned for the proposed metric learning algorithm. Furthermore, the experimental results imply that the hyperparameters introduced for automatic tuning of the distance threshold are insensitive to the accuracy.

\section*{Acknowledgments}
This work was supported by JSPS KAKENHI Grant Number
40401236. 

\bibliographystyle{plain}

\tslong{
  \appendix
\section{Proofs and Derivations}
\subsection{Proof of Lemma~\ref{lem:biasoptloss-is-bd}}
\label{ss:proof-lem:biasoptloss-is-bd}
We shall first show 
the equality \eqref{eq:Pb-is-bd}.
It suffices that both sides of \eqref{eq:Pb-is-bd}
has a term $\text{BD}(\cW,\cW_{0}\,;\,\varphi_{\text{r}})$.
Hence, it suffices to show the equality 
\begin{align}\label{eq:biasoptloss-is-bd}
  \text{Loss}(\vxi) + \constbregyo
  =
  \text{BD}(\vxi,\vxi_{0}\,;\,\varphi_{\text{l}})  
\end{align}
where we have defined 
\begin{equation}
\begin{aligned}
  & \varphi_{\text{l}}(\vxi) := \frac{1}{2}\left<\vxi,\vG\vxi\right>,
  \\
  & \text{Loss}(\vxi)
  :=
  \min_{b_{0}\in\bR}
  \widetilde{\text{Loss}}(\vxi,b_{0}) \qquad\text{with}
  \\
  & \widetilde{\text{Loss}}(\vxi,b_{0})
  :=
  \frac{c_{0}}{2}
  (b_{0}-\mu_{0})^{2}
  +
  \frac{c}{2}
  \sum_{k=1}^{K}(\xi_{k}-\gamma_{k}b_{0})^{2}.  
\end{aligned}
\end{equation}
The right hand side of \eqref{eq:biasoptloss-is-bd} is
\begin{align}
  \begin{split}
    \text{BD}_{\varphi_{\text{l}}}(\vxi,\vxi_{0})
  &=
  \varphi_{\text{l}}(\vxi) - \varphi_{\text{l}}(\vxi_{0})
  - \left<\nabla \varphi_{\text{l}}(\vxi_{0}), \vxi-\vxi_{0}\right>
  \\
  &=
  \frac{1}{2}\left<\vG\vxi,\vxi\right>
  -
  \frac{1}{2}\left<\vG\vxi_{0},\vxi_{0}\right>
  -
  \left<\vG\vxi_{0}, \vxi-\vxi_{0}\right>
  \\
  &=
  \frac{1}{2}
  \left<\vG\vxi, \vxi - 2\vxi_{0}\right>
  +
  \frac{1}{2}  
  \left<\vG\vxi_{0}, \vxi_{0}\right>. 
  \end{split}
\end{align}
Let us define 
\begin{align}\label{eq:optmu-for-biasoptloss}
  b_{0}(\vxi)
  :=
  \argmin_{b_{0}\in\bR}\widetilde{\text{Loss}}(\vxi,b_{0})
  =
  \mu_{1}\left<\vgam,\vxi\right> + \mu_{2}
\end{align}
to rewrite the loss function as
\begin{equation}
  \label{eq:21-prop:biasoptloss-is-bd}
  \begin{aligned}
    \text{Loss}(\vxi) &=
    \widetilde{\text{Loss}}(\vxi,b_{0}(\vxi))
    \\
    &=
    \frac{c}{2}\lVert\vxi-b_{0}(\vxi)\vgam\rVert^{2}
    +
    \frac{c_{0}}{2}
    \left(
      b_{0}(\vxi)-\mu_{0}
    \right)^{2}. 
  \end{aligned}
\end{equation}
See Sect.~\ref{ss:deriv-eq:optmu-for-biasoptloss}
for derivation of \eqref{eq:optmu-for-biasoptloss}. 
The above first and second terms, respectively, are
rearranged as
\begin{equation}\label{eq:1st-term-in-bdloss-proof}
\begin{aligned}
  \lVert\vxi-b_{0}(\vxi)\vgam\rVert^{2}
  &=
  \lVert\vxi-
  \left(
  \mu_{1}\left<\vgam,\vxi\right> + \mu_{2}
  \right)\vgam\rVert^{2}
  \\
  &=
  \lVert
  \left(
  \vI - \mu_{1}\vgam\vgam^\top
  \right)
  \vxi
  -
  \mu_{2}\vgam
  \rVert^{2}
  \\
  &=
  \mu_{2}^{2}\lVert\vgam\rVert^{2}
  +
  \left<\vxi,
  \vF^{2}
  \vxi
  -
  2\mu_{2}\vF\vgam
  \right>
\end{aligned}
\end{equation}
and
\begin{equation}\label{eq:2nd-term-in-bdloss-proof}
\begin{aligned}
  \left(
  b_{0}(\vxi)-\mu_{0}
  \right)^{2}
  &=
  \left(  
  \mu_{1}\left<\vgam,\vxi\right> 
  -\mu'_{0}
  \right)^{2}
  \\
  &=
  (\mu'_{0})^{2}
  +
  \left<\vxi,\mu_{1}^{2}\vgam\vgam^\top\vxi
  -2 \mu'_{0}\mu_{1}\vgam
  \right>, 
\end{aligned}
\end{equation}
where
\begin{equation}\label{eq:F-def-in-bdloss-proof}
\begin{aligned}
  \vF &:= \vI - \mu_{1}\vgam\vgam^\top, &
  \mu'_{0} &:= \mu_{0}-\mu_{2} = \mu_{1}\mu_{0}\lVert\vgam\rVert^{2}. 
\end{aligned}
\end{equation}
Substituting \eqref{eq:1st-term-in-bdloss-proof} and
\eqref{eq:2nd-term-in-bdloss-proof}
to
\eqref{eq:21-prop:biasoptloss-is-bd} and using
the equalities
\begin{equation}\label{eq:G-and-F-in-bdloss-proof}
  \begin{aligned}
    &\vG = c\vF^{2}+c_{0}\mu_{1}^{2}\vgam\vgam^\top, \qquad\text{and}
    \\
    &c\mu_{2}\vF\vgam + c_{0}\mu'_{0}\mu_{1}\vgam
    =
    \left(
    c \mu_{2} +
    \left(
    c_{0}\mu_{1}\mu_{0} - c\mu_{2}\right)
    \mu_{1}\lVert\vgam\rVert^{2}
    \right)
    \vgam, 
  \end{aligned}
\end{equation}
we have
\begin{equation}
\begin{aligned}
  \text{Loss}(\vxi)
  &=
  \frac{1}{2}
  \left(c \mu_{2}^{2}\lVert\vgam\rVert^{2} + c_{0}(\mu'_{0})^{2}\right)
  \\
  &\qquad\qquad
  +
  \frac{1}{2}
  \left<
  \vG\vxi,
  \vxi
  -2\vG^{-1}
  \left(
  c\mu_{2}\vF\vgam
  +
  c_{0}\mu'_{0}\mu_{1}\vgam
  \right)
  \right>
  \\
  &=
  \frac{1}{2}
  \left(c \mu_{2}^{2}\lVert\vgam\rVert^{2} + c_{0}(\mu'_{0})^{2}\right)
  -
  \frac{1}{2}
  \left<
  \vG\vxi_{0},\vxi_{0}\right>
  \\
  &\qquad\qquad\qquad
  +
  \frac{1}{2}
  \left<
  \vG\vxi,
  \vxi
  -2\vxi_{0}
  \right>
  +
  \frac{1}{2}
  \left<
  \vG\vxi_{0},\vxi_{0}\right>
  \\
  &=
  - \constbregyo + \text{BD}_{\varphi_{\text{l}}}(\vxi,\vxi_{0}), 
\end{aligned}
\end{equation}
which establishes 
the equality \eqref{eq:biasoptloss-is-bd}. 
Combining this result with the following lemma
concludes this proof. 
\begin{lemma}\label{lem:G-is-pd}
  In the setting of \eqref{eq:G-xi0-def-for-biasoptloss},
  $\vG$ is a strictly positive definite symmetric matrix
  if $c_{0}>0$, and
  $\vG$ is singular if $c_{0}=0$. 
\end{lemma}

See Sect.~\ref{ss:proof-of-lem:G-is-pd} for proof of
Lemma~\ref{lem:G-is-pd}. 

\subsection{Derivation of \eqref{eq:optmu-for-biasoptloss}}
\label{ss:deriv-eq:optmu-for-biasoptloss}
The derivative of
$\widetilde{\text{Loss}}(\vxi,b_{0})$
with respect to $b_{0}$ is given by
\begin{align}
  \nabla_{b_{0}}\widetilde{\text{Loss}}(\vxi,b_{0})
  =
  c_{0}(b_{0}-\mu_{0})
  +
  c \left<\vgam,\vgam b_{0} - \vxi\right>. 
\end{align}
Setting
$\nabla_{b_{0}}\widetilde{\text{Loss}}(\vxi,b_{0})=0$,
the stationary point is obtained as
\begin{align}
  b_{0}
  =
  \frac{c_{0}\mu_{0} + c\left<\vgam,\vxi\right>}
       {c_{0}+c\lVert\vgam\rVert^{2}}
  = \mu_{1}\left<\vgam,\vxi\right> + \mu_{2}
  = b_{0}(\vxi). 
\end{align}

\subsection{Proof of Lemma~\ref{lem:G-is-pd}}
\label{ss:proof-of-lem:G-is-pd}
Consider the case of $c_{0}>0$ first.
Here, we reuse the symbol $\vF$ defined in \eqref{eq:F-def-in-bdloss-proof}. 
We can observe that
\begin{align}
  \vG =
  c\vF^{2} + c_{0}\mu_{1}^{2}\vgam\vgam^\top
  \succ
  c\vF^{2}. 
\end{align}
Hence, it suffices to show 
$\vF^{2}\succ \vO$. 

Let us take an arbitrary vector $\vxi\in\bR^{K}\setminus\{\0\}$
and decompose the vector as
\begin{align}
  \vxi = \xi_{\parallel}\vgam + \vxi_{\perp}, 
\end{align}
where $\xi_{\parallel}\in\bR$ and
$\left<\vgam,\vxi_{\perp}\right>=0$.
From the assumption $\vxi\ne\0$,
either $\xi_{\parallel}\ne 0$ or
$\vxi_{\perp}\ne \0$ holds.  Then, we have
\begin{equation}
\begin{aligned}
  \vF\vxi &=
  (\vI-\mu_{1}\vgam\vgam^\top)
  (\xi_{\parallel}\vgam + \vxi_{\perp})
  =
  \left(
  1 - \mu_{1}\lVert\vgam\rVert^{2}
   \right)
   \xi_{\parallel}
  \vgam
  +
  \vxi_{\perp}
  \\
  &=
  \left(
  1-\frac{c\lVert\vgam\rVert^{2}}{c\lVert\vgam\rVert^{2}+c_{0}}
  \right)
   \xi_{\parallel}
  \vgam
  +
  \vxi_{\perp}
  =
  \frac{c_{0}\xi_{\parallel}}{c\lVert\vgam\rVert^{2}+c_{0}}
  \vgam
  +
  \vxi_{\perp}. 
\end{aligned}
\end{equation}
Therefore,
\begin{align}
  \left<\vxi,\vF^{2}\vxi\right>
  =
  \lVert\vF\vxi\rVert^{2}
  =
  \frac{c_{0}^{2}\lVert\vgam\rVert^{2}}{(c\lVert\vgam\rVert^{2}+c_{0})^{2}}
  \xi_{\parallel}^{2}
  +
  \lVert\vxi_{\perp}\rVert^{2}
  > 0
\end{align}
where the inequality follows from
the assumption that
either $\xi_{\parallel}\ne 0$ or
$\vxi_{\perp}\ne \0$ holds.

Next, let us discuss the case of $c_{0}=0$.
In this case, $\vG = c\vF^{2}$ and
the dimension of the \emph{kernel} of $\vF$ is non-zero because
\begin{align}
  \vF\vgam =
  (\vI-\mu_{1}\vgam\vgam^\top) \vgam
  =
  \frac{c_{0}}{c\lVert\vgam\rVert^{2}+c_{0}}
  \vgam
  =
  \0. 
\end{align}
This implies that $\vF$ is singular, and so is $\vG$. 

\subsection{Derivation of \eqref{eq:hk-order-K}}
From the definition of $\vG$,
it is easy to see that
$\vG$ is
the so-called \emph{rank-one update}
from the identity matrix,
which implies that the
inverse of $\vG$ can be expressed as 
\begin{align}
  \left[\vh_{1},\dots,\vh_{K}\right]
  = \vG^{-1} =
  \frac{1}{c}\left(\vI
  - \frac{\mu_{3}}{1+\mu_{3}\lVert\vgam\rVert^{2}}\vgam\vgam^\top\right). 
\end{align}
}

\end{document}